# Bangla-Bayanno: A 52K-Pair Bengali Visual Question Answering Dataset with LLM-Assisted Translation Refinement.

Mohammed Rakibul Hasan*[†], Rafi Majid*, Ahanaf Tahmid*
*Department of Electrical and Computer Engineering
North South University, Bashundhara, Dhaka-1229, Bangladesh
[†]Corresponding Author
Email: mohammed.hasan02@northsouth.edu

*Abstract*—In this paper, we introduce Bangla-Bayanno, an open-ended Visual Question Answering (VQA) Dataset in Bangla, a widely used, low-resource language in multimodal AI research. The majority of existing datasets are either manually annotated with an emphasis on a specific domain, query type, answer type, or are constrained by niche answer formats. In order to mitigate human-induced errors and guarantee lucidity, we implemented a multilingual LLM-assisted translation refinement pipeline. This dataset overcomes the issues of low-quality translations from multilingual sources. The dataset comprises 52,650 question–answer pairs across 4750+ images. Questions are classified into three distinct answer types: nominal (short descriptive), quantitative (numeric), and polar (yes/no). Bangla-Bayanno provides the most comprehensive open-source, high-quality VQA benchmark in Bangla, aiming to advance research in low-resource multimodal learning and facilitate the development of more inclusive AI systems.

## I. Introduction and Background

In recent years, multimodal AI has attracted significant attention from the research community and has become exceptionally popular. **Visual Question Answering (VQA)** is a fundamental multimodal task that demands AI systems to collaborate in interpreting natural language queries and visual content in order to provide accurate responses [1]. VQA has sparked significant attention in the fields of education, information retrieval, assistive technologies, and human-AI interaction. In order to facilitate the development of models that incorporate vision–language reasoning and enhance evaluation methodologies, benchmark datasets such as GQA [2], VizWiz [3], and VQA v2 [4] have been significant in the advancement of the field.

Visual Question Answering (VQA) research remains primarily focused on high-resource languages, notably English. Inclusivity and global applicability are significantly limited by this language inequality. Models that are predominantly trained on English data often struggle to generalise to linguistically diverse settings due to their lack of exposure to a variety of syntactic structures, cultural contexts, and linguistic nuances. The development of robust multimodal AI systems is impeded by the scarcity of high-quality annotated datasets and the lack of reliability of direct machine translations [5] for low-resource languages.

This disparity can be seen by Bengali, which is the seventh most spoken language in the world and is spoken by more than 230 million people. Despite this, Bangla is under-represented in multimodal AI research. The development and evaluation of Bangla-specific VQA models are impeded by the limited extent and quality of the existing resources.

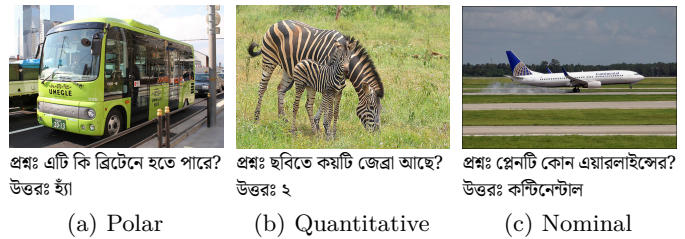

(a) Polar  (b) Quantitative  (c) Nominal

Fig. 1: Examples from the **Bangla-Bayanno** dataset. Each instance includes an image, a corresponding question in Bangla, and an answer. The examples represent the three distinct answer types: (a) Polar (yes/no), (b) Quantitative (numeric), and (c) Nominal (descriptive).

Bangla-Bayanno is the most extensive open-source VQA benchmark in Bangla, comprising three distinct QA categories, is released to address this challenge. Its objective is to facilitate the development of AI systems that are globally applicable and linguistically inclusive, as well as to advance research in low-resource multimodal learning.

Developing a high-quality Bangla VQA dataset is a challenging endeavour because of translation constraints. Traditional manual annotation methods, such as Excel-style entry, frequently introduce inconsistencies and unnatural grammar. On the other hand, generic machine translation (MT) tools produce grammatically accurate but semantically superficial translations, often overlooking contextual nuances that are unique to Bangla. In order to circumvent these constraints, we developed a multilingual refinement pipeline that is assisted by LLM and that maintains semantic clarity and generates fluent, contextually accurate translations [6]. This approach ensures that Bangla-Bayanno is a sustainable benchmark for low-resource multimodal AI research, as it not only scales to thousands of question–answer pairs but also maintains linguistic integrity.

| Sentence (English) | Manual (Excel-style) | Baseline MT Tool | Bangla-Bayanno |
|---|---|---|---|
| What are the vases sitting on? | ফুলদানি কি বসে আছে? | ফুলদানি কি বসে আছে? | ফুলদানিগুলি কীসের উপর রাখা আছে? |
| How many blue players are shown? | কত নীল খেলোয়াড় দেখানো হয়? | কত নীল খেলোয়াড় দেখানো হয়? | কয়জন নীল দলের খেলোয়াড় দেখানো হয়েছে? |
| Is this location in America? | এই অবস্থানটি কি আমেরিকায়? | আমেরিকাতে এই অবস্থান কি? | এই স্থানটি কি আমেরিকায়? |
| How many rackets are there? | কত র্যাকেট আছে? | কত র্যাকেট আছে? | কয়টি রেকেট আছে? |

TABLE I: Comparison of translations across methods.

Table I illustrates the quality of translations across various methodologies through color-coded indicators. Red signifies outputs that are either partially correct or defective, resulting in confusing or only partially retained meaning. Such concerns often arise in manual (Excel-style) translations, characterized by code-mixing or literal wording, as well as in baseline machine translation technologies, which frequently yield superficial or contextually unsuitable translations. In contrast, green indicates translations that are entirely accurate, fluent, and contextually appropriate, as routinely produced by the Bangla-Bayanno pipeline. This visual differentiation clearly illustrates that Bangla-Bayanno mitigates ambiguity and errors, providing high-quality, legible translations appropriate for low-resource multilingual benchmarking.

## II. Related Work
### A. Existing Datasets and Low-Resource Languages

The provision of diverse benchmarks has significantly advanced vision–language reasoning. The "language prior" problem was mitigated by VQA v2 [4], which balanced complementary image–question pairings to ground reasoning in the visual domain. GQA [2] advanced compositional reasoning through scene-graph–based questions and rigorous metrics, while VizWiz [3] introduced real-world challenges with unanswerable, conversational queries and low-quality images from blind users. Although these datasets collectively propelled progress, their impact has been disproportionately concentrated in English.

This restricts inclusivity and limits multimodal AI for a global audience. Despite recent NMT advances in low-resource African languages [5], progress depends on focused data and benchmarks. The absence of such resources for Bangla—a language with a vast speaker base—remains a significant impediment. Recent initiatives include *BengaliVQA* (13k balanced yes/no pairs over 3k images) [7], *ChitroJera* (15k culturally relevant QA pairs) [8], and *Med-VQA_Bn_Overall* (7k medical entries over 600+ categories) [9]. LLM-driven datasets like *BVQA* (17.8k open-ended pairs across 3.5k images) [10] and *VQA Bengali 1.0* (3.7k yes/no pairs via contrastive loss) [11] extend coverage, though existing datasets remain limited in scale, answer variety, or linguistic depth.

### B. Conventional Translation and Annotation Approaches

In low-resource contexts, researchers frequently employ machine translation (MT) [12] to expand VQA datasets, though outputs are often literal and lack context [13]. MT effectiveness is restricted by sparse parallel data, domain mismatches, and noisy corpora, while NMT continues to underperform in comparison to high-resource settings [14].

Beyond MT, Bangla resources often use manual "Excel-style" workflows for translation or annotation. For example, BanglaTense applied Google Sheets for tense labeling [15], BanglaBlend distributed English–Bangla parallels in Excel [16], and prior MT corpora aligned text similarly [17]. These methods are labor-intensive, annotator-dependent, and insufficient for scaling multimodal datasets, despite claims of higher linguistic quality.

### C. LLM-Assisted Translation Refinement

Recently, large language models (LLMs) such as GPT-3.5, GPT-4, and ChatGPT [18]–[20] have shown remarkable performance in translation tasks, even for low-resource languages as Bangla [18], [20]. In contrast to traditional machine translation (MT) systems, these models employ contextual reasoning and extensive pretraining to produce fluent, contextually accurate translations, reduce literal or code-mixed outputs, and resolve syntactic ambiguity. LLMs outperform traditional neural MT systems in numerous low-resource scenarios by creating translations that maintain cultural context and semantic nuance, as established by recent evaluations [21], [22]. In addition, their API-based accessibility [20] permits the scalable integration of their services into multilingual NLP pipelines wherever diverse corpora are unusual.

In this work, we utilize the generative refinement capabilities of ChatGPT through API-driven workflows to enhance the quality of translations. This refinement layer systematically resolves syntactic awkwardness and semantic ambiguity in Bangla queries and answers, resulting in translations that are more precise, scalable, and fluent across thousands of instances. The adaptability of LLMs as practical tools for constructing robust, high-quality multilingual multimodal datasets is further demonstrated by this integration.

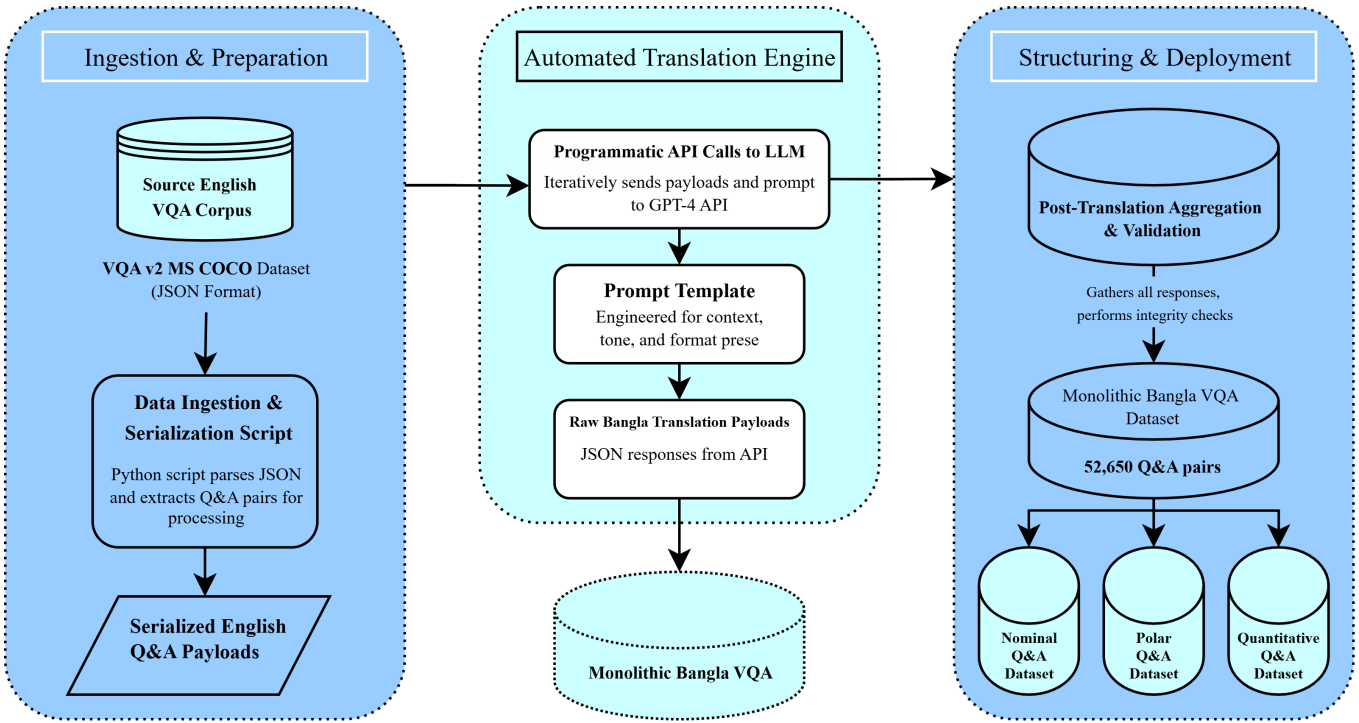

Fig. 2: Workflow diagram of the Bangla-Bayanno.

## III. METHODOLOGY

The Bangla-Bayanno pipeline leverages an LLM-assisted refinement process to provide high-quality, context-preserving translations for low-resource settings, as shown in Fig. 2. In order to assure syntactic consistency, accuracy, and scalable dataset construction, the process is divided into three sequential phases: Ingestion and Preparation, Automated Translation, and Structuring and Deployment. A corpus of 52,650 question-answer pairings over 4,750+ images, categorized into nominal, polar, and quantitative categories, was generated by this process, minimizing human-induced errors.

### A. Ingestion and Preparation

*1) Data Collection:* - Bangla-Bayanno is based on the VQA v2 dataset [1], [4], which is the most extensively utilized benchmark for multimodal reasoning. Developed on the MS COCO dataset, it has 265,000 images and 1.1 million question-answer pairs, encompassing a variety of reasoning tasks, including recognition, counting, and reasoning. In contrast to VQA v1, VQA v2 addresses language bias by associating each question with corresponding visuals, compelling models to depend on both visual and linguistic inputs [4]. The scale, diversity, and equitable design of VQA v2 have recognized it as the standard resource in VQA research [23].

For Bangla-Bayanno, we randomly selected around 4,750 images along with their corresponding question-answer pairs from this dataset, ensuring a diverse visual representation and contextually relevant annotations. This foundation enables us to apply the rigor of a globally acknowledged benchmark to the Bangla language, hence overcoming the low-resource gap.

*2) Data Processing:* - Bangla-Bayanno was curated through a systematic preprocessing pipeline to ensure consistency and reproducibility. Question–answer pairs were extracted and aligned with their corresponding images.

```
{
    "qa_id": 0,
    "image_id": 480056,
    "image_file": "COCO_train2014_000000480056.jpg",
    "question_en": "Is this animal in the wild?",
    "answer_en": "no"
}
```

Fig. 3: Example JSON schema from the dataset.

Each entry was structured into a unified JSON

schema as Fig. 3, embedding the QA identifier, image reference, question text, answer text, and categorical answer type (polar, numeric, descriptive). This serialized standard format guarantees interoperability and facilitates downstream multimodal tasks. It was confirmed that every question–answer pair is correctly linked to an image file, while statistical validation ensured balanced coverage across the three answer categories.

### B. Automated Translation

*1) Utilization of Large Language Model:* We utilized Microsoft Azure's deployment of ChatGPT-4. This model has exhibited enhanced translation proficiency relative to alternative systems [24]. Conducted a thorough assessment revealing that GPT-4's translation capabilities are on par with those of junior-level human translators across several languages and areas, including Bengali. GPT-4 demonstrates constant performance across many language pairs, including low-resource languages like Bengali, without substantial deterioration in translation quality.

Fig. 4: Prompt template used in the Bangla-Bayanno.

*2) Prompts Engineering:* The acquired English VQA corpus is subsequently utilized to translate, refine, and enhance the dataset in Bengali. Each entry is serialized into structured JSON and delivered to the LLM using Azure, following a well-crafted prompt. This procedure guarantees precise translation, contextual development, and the generation of a transformed dataset in Bengali while maintaining the original structure. The prompt template that is used is shown in Fig. 4.

Fig. 5: Example JSON schema after translation.

*3) Monolithic Bangla Question-Answering Dataset:* Following to the preliminary translation phase, each question-answer pair endured to a large language model (LLM) to guarantee syntactic fluency, semantic precision, and contextual coherence in Bangla as shown in Fig. 5 . To maintain robustness, translations were conducted individually, hence minimizing group noise effects sometimes associated with batch translation processes. An automated computer managed this pipeline by sequentially querying the LLM with distinct entries and preserving the generated Bangla outputs. This modular loop facilitated uniform processing of nominal, quantitative, and polar categories, while enabling the model to address unclear phrasing, unnatural syntax, or literal machine translation problems. All improved responses were compiled into a unified, comprehensive dataset in Bangla. This dataset exhibits enhanced structural consistency and readability, providing a linguistically accurate and semantically strong standard. The refinement approach guarantees that Bangla-Bayanno is not simply a translated corpus but a high-quality, domain-adapted resource tailored for low-resource multimodal learning.

### C. Structure and Deployment

*1) Aggregation and Validation:* The post-translation phase included automated consistency verifications and focused manual evaluations to guarantee dataset dependability. Minor inaccuracies, including occasional incorrect interpretation of numerical values, were rigorously rectified. This procedure maintained structural integrity and semantic accuracy while reducing human bias. Consequently, the final Bangla VQA dataset achieved linguistic precision and contextual robustness, establishing it as a reliable research standard.

| Answer Type | QA Pairs | Images | Access |
|---|---:|---:|---|
| Polar (Yes/No) | 19,619 | 4,163 | Config: polar |
| Numeric | 6,484 | 2,517 | Config: numeric |
| Descriptive | 26,033 | 4,451 | Config: descriptive |
| **Full** | **52,650** | **4,750+** | Config: Full dataset |

TABLE II: Bangla-Bayanno release variants and coverage.

*2) Categorization and Deployment:* The final Bangla-Bayanno dataset comprises **52,650** question–answer (QA) pairs across **4,750+** images. Each QA instance is automatically categorized into one of three answer types, *polar* (yes/no), *quantitative* (numeric), or *descriptive* (short textual) to support targeted benchmarking and analysis. We release four access-friendly variants on the Hugging Face Hub: the three category-specific subsets and the complete corpus, visible on Table II. Public hosting facilitates reproducibility and easy integration into research pipelines via standard Hub APIs.[1]

By providing this structured categorization and modular access, Bangla-Bayanno enables the community to pursue task-specific VQA evaluation while also supporting holistic

---
[1]Dataset URL: `Remian9080/Bangla-Bayanno`.

multimodal benchmarking in Bangla, a low-resource yet globally significant language.

## IV. Dataset Analysis and Evaluation

### A. Statistics & Distribution

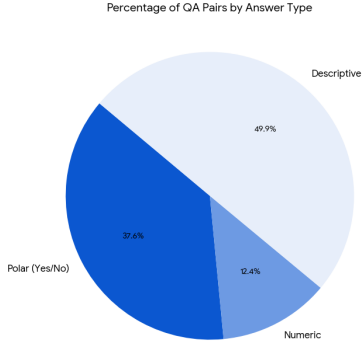

Fig. 6: Percentage of QA Pairs by Answer Type.

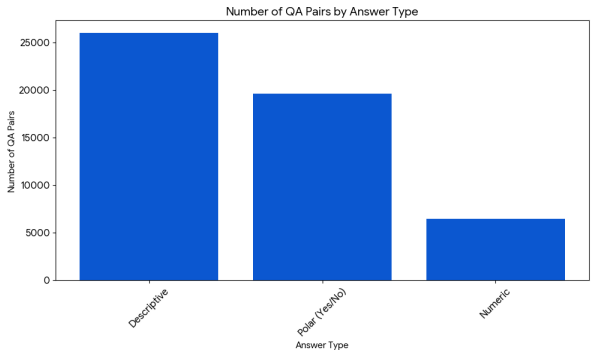

Fig. 7: Number of QA Pairs by Answer Type.

The dataset analysis indicates a clear compositional structure, as outlined in Fig. 6 and Fig. 7 and illustrated in the corresponding bar and pie charts. The dataset consists of Nominal questions, totaling 26,033 pairings (49.9% of the whole), followed by Polar questions at 19,619 pairs (37.6%), and concluding with Quantitative questions at 6,484 pairs (12.4%). Subsequent examination of the token indicates that Nominal responses are comparatively lengthier, whereas Polar responses are the most concise, generally including a single token. This distribution indicates that the dataset has a greater emphasis on qualitative and binary reasoning tasks, with Nominal and Polar questions comprising over 87% of the data.
An automated analysis responded that the dataset also exhibits succinct language patterns, with an average question length of 6.21 tokens and an average answer length of 1.09 tokens, reflecting the predominance of short, focused responses, also a factor to choose Bangla-Bayanno, while working with focused environment.

### B. Translation Quality Evaluation

| Score | Description |
|---|---|
| 1 | Completely incorrect; broken or meaningless output. |
| 3 | Partially correct; some words untranslated, meaning understandable. |
| 5 | Correct translation; meaning intact with fluent syntax. |

TABLE III: Translation Quality Evaluation Criteria

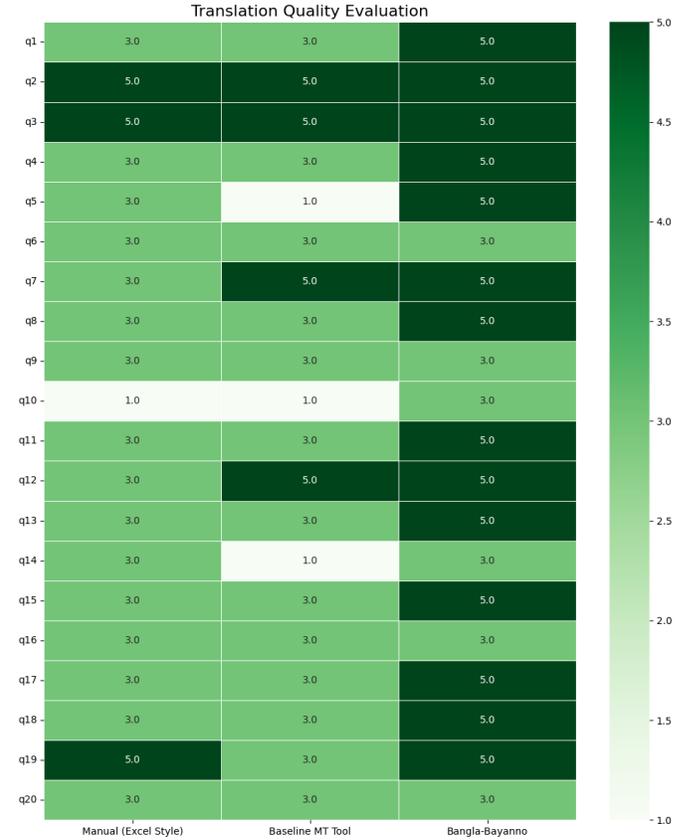

TABLE IV: Translation quality evaluation across 20 questions using Heatmap

The findings from 20 example questions (q1–q20) shown in the Fig. IV explains that the Bangla-Bayanno approach frequently achieves better ratings based on Table III, often attaining the maximum rating of 5, thereby surpassing both manual (Excel-style) translations and the baseline MT tool, while the manual translations are generally clear (predominantly rated 3), the baseline MT tool displays varying results with rare failures (scores as low as 1), Bangla-Bayanno showcases superior semantic precision and syntactic fluency, underscoring its efficacy in generating high-quality Bangla translations.
The evaluation involved iterative assessments of each translation by multiple group members, with the final score determined by calculating the mean of their ratings. The evaluators, who graduated from different fields, were well-versed in Bangla and English.

## C. Comparative Analysis

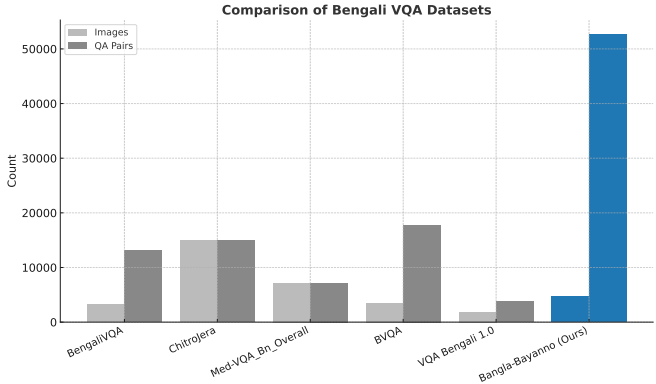

Fig. 8: Comparison of Bengali VQA Datasets by the number of Images and Question-Answer (QA) Pairs. The chart visually demonstrates that Bangla-Bayanno contains a substantially larger number of QA pairs than all other listed datasets.

| Dataset | Images | QA Pairs | Answer Types |
|---|---|---|---|
| BengaliVQA | 3,280 | 13,046 | Yes/No |
| ChitroJera | 15,000+ | 15,000+ | Mixed |
| BVQA | 3,500 | 17,800 | Open-ended |
| VQA Bengali 1.0 | 1,864 | 3,728 | Yes/No |
| Med-VQA | 7,170 | 7,170 | Medical-specific |
| **Bangla-Bayanno** | **4,750+** | **52,650** | Nominal, Quant., Polar |

TABLE V: Bengali VQA datasets and Bangla-Bayanno

Figure 8 visually compares the scale of existing Bengali VQA datasets, which are summarized alongside Bangla-Bayanno in Table V. Early efforts such as *BengaliVQA* [7] and *VQA Bengali 1.0* [11] are limited to binary yes/no questions, restricting their ability to capture diverse reasoning patterns. *ChitroJera* [8] expands the scope with culturally relevant pairs but remains relatively modest in QA scale. Domain-specific initiatives like *Med-VQA_Bn_Overall* [9] provide valuable medical coverage yet are narrowly focused and smaller in size. More recent LLM-driven datasets such as *BVQA* [10] increase openness in question formulation but depend primarily on synthetic generation and cover fewer images overall.

In contrast, Bangla-Bayanno surpasses these efforts by providing over 52,650 question–answer pairs across more than 4,750 images, covering nominal, quantitative, and polar categories. Its multilingual LLM-assisted refinement pipeline ensures linguistic accuracy and reduces translation-induced artifacts, thereby addressing both scale and quality limitations simultaneously. Bangla-Bayanno thus establishes the most comprehensive, diverse, and linguistically robust open-source VQA benchmark in Bangla to date.

## V. Computational Cost

The anticipated computational expense for the AI refinement pipeline utilized in the development of the Bangla-Bayanno dataset is estimated to range from $110.57 to $131.63. This computation involves 52,650 API calls to the Azure OpenAI Service, employing the GPT-4o model. The estimate predicts a consumption rate of 180-200 input tokens and 80-100 output tokens per invocation. This cost study utilizes the official Azure pricing for the designated model, set at $0.005 per 1,000 input tokens and $0.015 per 1,000 output tokens, thereby assessing the resources necessary for the data refining process.

## VI. Conclusion and future work

We introduced Bangla-Bayanno, an extensive, open-ended Visual Question Answering dataset in Bangla, comprising 52,650 question-answer pairings across over 4,750+ images. The dataset employs a multilingual LLM-assisted workflow to reduce annotation errors and enhance translation quality, providing the most extensive VQA benchmark for a low-resource language.

Compared to prior datasets such as BengaliVQA, ChitroJera, BVQA, VQA Bengali 1.0, and Med-VQA, Bangla-Bayanno is not only larger in scale but also richer in answer-type diversity (nominal, quantitative, and polar), positioning it as a new benchmark for low-resource VQA research.

However, some translations preserve source-language terminology without complete modification, and the visual scope may be further enhanced. Future endeavors will concentrate on (i) human-in-the-loop validation to guarantee linguistic accuracy, (ii) augmenting the dataset with a greater variety of images, and (iii) expanding it to encompass cross-lingual visual question answering to facilitate extensive multimodal research in low-resource environments.